%% file: main.tex
\pdfoutput=1 
\documentclass[letterpaper, 10 pt, journal, twoside]{IEEEtran}

\IEEEoverridecommandlockouts                              

\usepackage{times} 
\usepackage{amsmath} 
\usepackage{amssymb}  
\usepackage{url}
\usepackage{latexsym}
\usepackage{graphicx}	
\usepackage{comment}
\usepackage{threeparttable}
\usepackage{scalefnt}
\usepackage{multicol}
\usepackage{wrapfig}
\usepackage{algorithm} 
\usepackage{algorithmic} 
\usepackage{here} 
\usepackage{autobreak} 
\usepackage{breqn}
\usepackage{color}

\makeatletter  
\newcommand\fs@spaceruled{\def\@fs@cfont{\bfseries}\let\@fs@capt\floatc@ruled
  \def\@fs@pre{\vspace{0.5\baselineskip}\hrule height.8pt depth0pt \kern2pt}%
  \def\@fs@post{\kern2pt\hrule\relax}%
  \def\@fs@mid{\kern2pt\hrule\kern2pt}%
  \let\@fs@iftopcapt\iftrue}
\makeatother

\title{Embodying pre-trained word embeddings \\through robot actions}

\author{Minori Toyoda$^{1}$, Kanata Suzuki$^{1,2}$, Hiroki Mori$^{3}$, Yoshihiko Hayashi$^{1}$ and Tetsuya Ogata$^{1,4}$
\thanks{Manuscript received: October, 15, 2020; Revised January, 19, 2021; Accepted February, 17, 2021.}
\thanks{This paper was recommended for publication by Editor Tamim Asfour upon evaluation of the Associate Editor and Reviewers’ comments.
This work was supported by JST CREST Grant Number JPMJCR15E3, Japan.} 
\thanks{$^{1}$Minori Toyoda is with Faculty of Science and Engineering, Waseda University, Tokyo 169-8555, Japan,
    {\tt\footnotesize minori-toyoda@fuji.waseda.jp}.}
\thanks{$^{1,2}$Kanata Suzuki is with Artificial Intelligence Laboratories, Fujitsu Laboratories LTD., Kanagawa 211-8588, Japan,
and also with Faculty of Science and Engineering, Waseda University, Tokyo, Japan,
    {\tt\footnotesize suzuki.kanata@fujitsu.com}.}
\thanks{$^{3}$Hiroki Mori is with Institute for AI and Robotics, Future Robotics Organization, Waseda University, Tokyo 169-8555, Japan,
    {\tt\footnotesize mori@idr.ias.sci.waseda.ac.jp}.}
\thanks{$^{1}$Yoshihiko Hayashi is with Faculty of Science and Engineering, Waseda University, Tokyo, Japan,
    {\tt\footnotesize yshk.hayashi@aoni.waseda.jp}.}
\thanks{$^{1,4}$Tetsuya Ogata is with Faculty of Science and Engineering, Waseda University, Tokyo, Japan,
and also with National Institute of Advanced Industrial Science and Technology, Tokyo 100-8921, Japan,
    {\tt\footnotesize ogata@waseda.jp}.}
\thanks{Digital Object Identifier (DOI): see top of this page.}
}

\begin{document}

\markboth{IEEE Robotics and Automation Letters. Preprint Version. Accepted February, 2021}
{Toyoda \MakeLowercase{\textit{et al.}}: Embodying pre-trained word embeddings through robot actions}

\maketitle

\begin{abstract}
We propose a promising neural network model with which to acquire a grounded representation of robot actions and the linguistic descriptions thereof. 
Properly responding to various linguistic expressions, including polysemous words, is an important ability for robots that interact with people via linguistic dialogue. 
Previous studies have shown that robots can use words that are not included in the action-description paired datasets by using pre-trained word embeddings. 
However, the word embeddings trained under the distributional hypothesis are not grounded, as they are derived purely from a text corpus.
In this paper, we transform the pre-trained word embeddings to embodied ones by using the robot's sensory-motor experiences. 
We extend a bidirectional translation model for actions and descriptions by incorporating non-linear layers that retrofit the word embeddings. 
By training the retrofit layer and the bidirectional translation model alternately, our proposed model is able to transform the pre-trained word embeddings to adapt to a paired action-description dataset. 
Our results demonstrate that the embeddings of synonyms form a semantic cluster by reflecting the experiences (actions and environments) of a robot.
These embeddings allow the robot to properly generate actions from $\it unseen$ $\it words$ that are not paired with actions in a dataset.
\end{abstract}


\input{subtex/0intro}
\input{subtex/1related}

\input{subtex/2model}
\input{subtex/3experiment}

\input{subtex/4result}
\input{subtex/5conclusion}

\input{main.bbl}
\end{document}

%% file: subtex/0intro.tex
\section{Introduction}

\IEEEPARstart{T}{here} has recently been a growing interest in robots that can interact with people using linguistic dialogue. 
Such robots must be able to flexibly generate actions based on linguistic instructions and verbally explain their actions or situations around them~\cite{yamada2018paired}.
To achieve this requirement, it is important to appropriately associate various linguistic expressions, including polysemous words, with actions and environments. 
Deep neural networks (DNNs) are an effective approach to acquire an integrated representation of such expressions. 
However, it is difficult to prepare a large training dataset that includes robot actions, information on the environment, and the descriptions with a large variety.
Robots need to ground even ``unseen words" to the real world with limited data.
Here, we define ``unseen words" as \textit{words that have pre-trained embeddings but are not associated with robot actions.}

Some approaches currently in operation use pre-trained word embeddings to generate robot actions~\cite{lynch2020grounding}\cite{matthews2019word2vec}.
However, these word embeddings are trained under the distributional hypothesis~\cite{harris1954distributional} and are not suitable for generating actions.
That is, a word embedding
acquires similar representations for words that appear in similar contexts in the text corpus, even if they are antonyms such as ``fast" and ``slowly." 
This may lead to a situation, where robots generate unexpected behavior for unseen words.

\begin{figure}[tb]
    \centering
    \includegraphics[width=8.6cm]{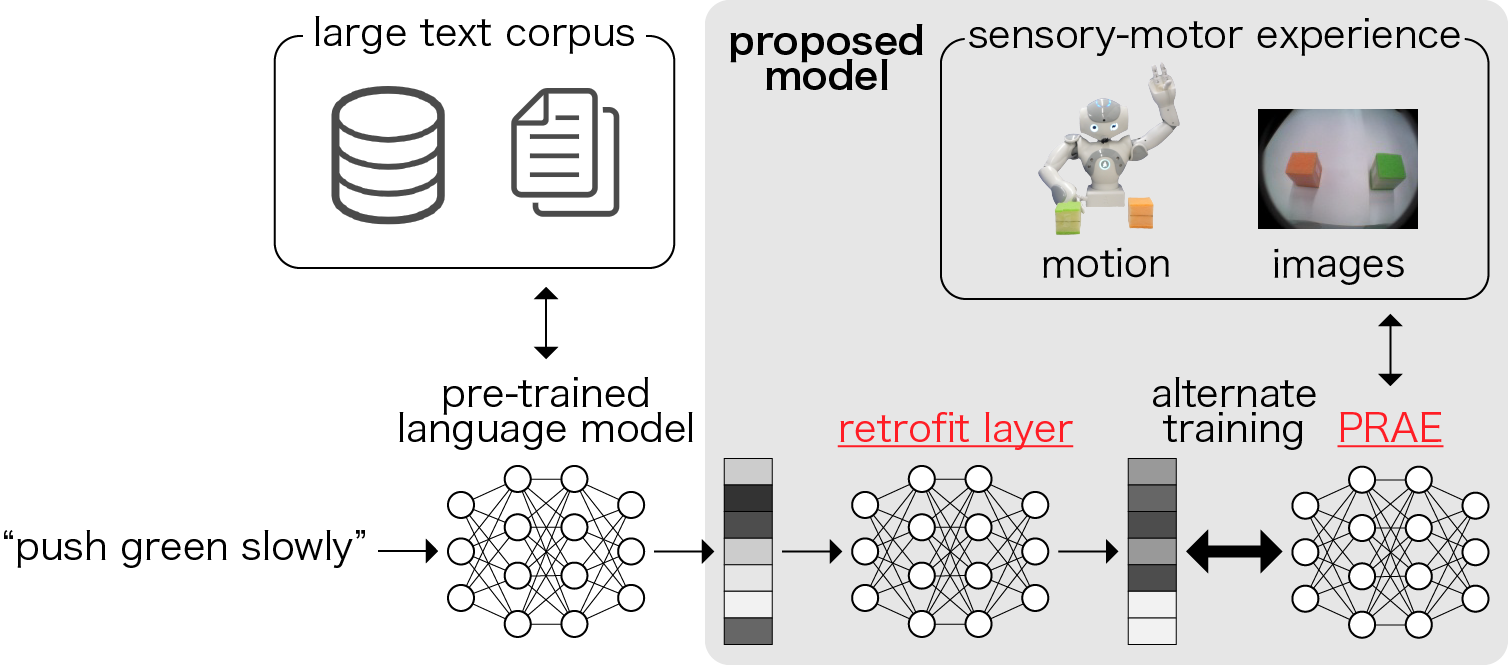}
    \caption{
       Our proposed framework transforms word embeddings by observing robot actions and generates appropriate actions for unseen words using the retrofit layer and the Paired Recurrent Autoencoder (PRAE).
    }
 \label{fig:trans_task}
\end{figure}

The purpose of this study is to develop a model that can translate between actions and linguistic descriptions even if the latter contains unseen words.
To this end, we propose a neural framework as illustrated in Fig.~\ref{fig:trans_task}.
This framework, abbreviated as rPRAE (retrofit Paired Recurrent Autoencoder), can refine pre-trained word embeddings tailored to a robot’s environment by employing the \textit{retrofit} layer.
Refined, or \textit{grounded}, word embeddings are able to address the innate issue of word embedding trained through a purely linguistic process.

The proposed rPRAE model is a seq2seq-based model that bidirectionally translates between actions and their descriptions.
By incorporating the dedicated retrofit layer, its architecture is extended from that of the Paired Recurrent Autoencoder (PRAE) model proposed by Yamada et al.~\cite{yamada2018paired}.
This layer is arranged between pre-trained word embeddings and the PRAE such that it refines embeddings that are tailored to the robot's sensory–motor experiences.

Here, ``retrofit" refers to the transformation of pre-trained embeddings acquired under the distributional hypothesis into a vector representation compatible with external knowledge structures and information from non-text modalities~\cite{faruqui2014retrofitting}.
We also propose to train the retrofit layer and the PRAE alternately, which enables the conversion of pre-trained word embeddings into representations adapted to the actions and environments of the robot, while preventing overfitting.

The contributions of the present work to the field in general can be summarized as follows:
\begin{itemize}
    \item We proposed a model to bidirectionally translate between robot actions and linguistic descriptions that may include unseen words by retrofitting pre-trained word embeddings.
    \item The model acquires grounded representations that cannot be obtained by the distributional hypothesis.
\end{itemize}

%% file: subtex/1related.tex
\section{Related Work}

Many studies into verbal communication with robots have been conducted, where the main challenge is to associate a sequence of words, i.e., a linguistic description, with a sequence of physical movements, i.e., a robot action~\cite{yamada2018paired}\cite{lynch2020grounding}\cite{matthews2019word2vec}\cite{ahn2018text2action}\cite{ogata2007two-way}\cite{tuci2011experiment}\cite{juven:hal-02594725}\cite{sugita2005learning}\cite{zhong2017understanding}.
In light of this challenge, several attempts have been made to explore better representations of linguistic units, as well as efficient yet effective computational mechanisms~\cite{FODOR19883}\cite{DBLP:conf/icml/LakeB18} enabling some level of compositionality to be achieved.
In the remainder of this section, we briefly review two research areas that are vital in achieving this goal.

\subsection{Action Generation using Pre-trained Word Embeddings}

The majority of studies that combine descriptions and actions used one-hot vectors or vectors acquired in experiments to represent linguistic units~\cite{yamada2018paired}\cite{ogata2007two-way}\cite{tuci2011experiment}\cite{sugita2005learning}\cite{lin2018generating}\cite{suzuki2018ral}.  
However, unlike one-hot vectors, using pre-trained word embeddings derived from a large corpus generally permits the measurement of the similarity or relatedness between words. 
This characteristic provides an ability to manage ``unseen words" that do not appear in the paired dataset. 

Some studies generated actions from descriptions represented by pre-trained word embeddings~\cite{lynch2020grounding}\cite{matthews2019word2vec}\cite{ahn2018text2action}\cite{zhong2017understanding}\cite{ahuja2019language2pose}. 
Zhong et al.~\cite{zhong2017understanding}, Matthews et al.~\cite{matthews2019word2vec}, and Lunch et al.~\cite{lynch2020grounding}, in particular, generated actions from commands including unseen words. 
Zhong et al.~\cite{zhong2017understanding} used averaged pre-trained word embeddings for this task; however, this approach fails to generate actions from descriptions whose meaning depends on the order of the words.
Although Matthews et al.~\cite{matthews2019word2vec} generated actions from single-word commands, the correspondence between a command and the corresponding action is fixed.

These studies used pre-trained embeddings without fine-tuning them to be integrated with actions. 
Pre-trained embeddings of action concepts acquired from a linguistic corpus are not suitable for generating actions in a real environment as they may result in inappropriate actions.
The present work resolves this issue by employing retrofitted word embeddings that maintain the associations between a word meaning and the action and environment of the robot.
This differs from previous work relying on cross-situational learning~\cite{juven:hal-02594725}, which focuses on resucing supervision signals; this could be the basis for future studies.

\subsection{Multi-modal Representations of Semantics}

The distributed representations of words, also known as ``word embeddings," are employed in almost every sub-field of natural language processing, as demonstrated by Word2Vec~\cite{mikolov2013distributed}.
These resources are built from a corpus of written texts by relying on the distributional hypothesis that argues that ``words that occur in the same contexts tend to have similar meanings"~\cite{Firth1957}.
This approach enables the representation of words in vocabulary using low-dimensional dense vectors.

Although the distributional hypothesis provides a solid foundation for mathematically-sound representations of words, it does have some innate issues~\cite{hasegawa-etal-2018-social}\cite{doi:10.1111/tops.12093}.
For example, since it focuses on the contextual similarity of a word, words that frequently appear in similar contexts cannot be clearly distinguished by the vector representations; therefore, semantically coordinated words and antonyms tend to have similar vectors.
Another issue is that the representations cannot capture semantics that are not explicitly written in the corpus.
For example, human perceptual experience or knowledge based on common sense is hardly ever extracted from the ordinary corpus.

To address these issues, several attempts have been made to integrate information obtained from media other than language with the embeddings pre-trained with a text corpus, but most of the work concentrates on nominal concepts and the representation of verbs remains a challenging problem~\cite{beinborn-etal-2018-multimodal}.
The present work acquires grounded representations that are necessary for dealing with robot actions.
It also demonstrates that our research setting would enable the study of semantic representations of verbs and adverbs grounded in a physical environment.

%% file: subtex/2model.tex
\section{Method}

\begin{figure*}[htb]
    \vspace*{1.5pt}
    \centering
    \includegraphics[width=17.8cm]{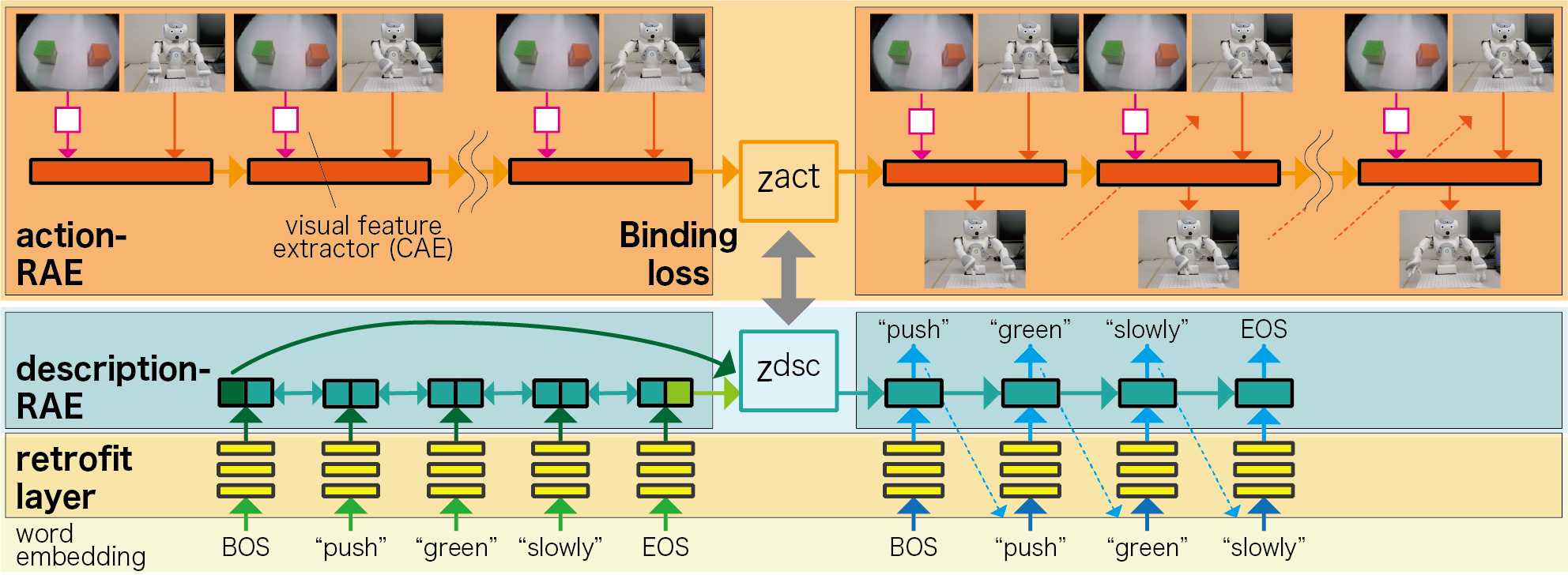}
    \caption{
        Model overview. 
        The proposed model consists of two recurrent autoencoders and a retrofit layer.
    }
    \label{fig:model}
    \vspace*{-2pt}
\end{figure*}

This study proposes a novel neural network model with which to acquire grounded representations of robot actions and their linguistic descriptions. 
As shown in Fig.~\ref{fig:model}, the proposed model consists of two recurrent autoencoders (RAEs) for descriptions and the robot’s sensory-motor experience as well as a retrofit layer to convert pre-trained word embeddings. 
In this study, we define robot actions as sequences of joint angle values, and visual information as RGB images obtained from the head camera on the robot. 
We also define descriptions as word sequences.

\subsection{Paired Recurrent Autoencoders}

We adopted a paired recurrent autoencoder (PRAE) as the learning model to integrate representations of linguistic descriptions and robot actions~\cite{yamada2018paired}.
The PRAE enables robots to both produce actions in response to descriptions and produce descriptions given their own actions. 
The PRAE consists of two RAEs; one for robot action sequences and one for word sequences. 
We train each RAE to minimize the reconstruction error of the original sequences. 
In addition, the model has an additional loss function that forces the representation of actions and their descriptions to be close to each other in a vector space.

In the description-RAE, the encoder recurrent neural network (RNN) embeds a word sequence of length $T_d$ into vector $z^{dsc}$ and the decoder RNN generates a sequence from $z^{dsc}$. 
The formulation of the reconstruction loss of description-RAE is as follows:
\vspace{-2pt}  
\begin{eqnarray}
    L_{\rm dsc} = \frac{1}{T_d-1} \sum _{t=1}^{T_d-1} \left( - \sum _w^W \hat{x}_{t+1} (w) \log y_t (w) \right).
\end{eqnarray}
Here, $W$ is the vocabulary size, $\hat{x}$ is a one-hot vector that specifies the target words, 
and $y$ is the output value activated through a softmax function.
For the input of the description-RAE, we use word embeddings through the retrofit layer .
If unseen words are inputted, a proper vector $z^{dsc}$ may not be obtained in unidirectional encoding depending on the word order.
To mitigate this issue, we use a bidirectional long short-term memory (biLSTM)~\cite{graves2005framewise}.

In the action-RAE, we input joint angles and image features that are pre-compressed in a convolutional autoencoder (CAE)~\cite{masci2011stacked} to the model.
The encoder RNN embeds the robot action sequence of length $T_a$ into vector $z^{act}$, and the decoder RNN generates only the joint angle sequences from $z^{act}$.
The reconstruction loss of the action-RAE is a mean squared error as shown below:
\vspace{-1pt}  
\begin{eqnarray}
    L_{\rm act} = \frac{1}{T_a-1} \sum _{t=1}^{T_a-1} || j_{t+1} - \hat{j}_{t+1} ||_2^2.
\end{eqnarray}
\vspace{-1pt}  
\newline  
Here, $j$ and $\hat{j}$ dictate the target value of the joint angles and the predicted value of the joint angles, respectively.

In addition to the reconstruction losses $L_{dsc}$ and $L_{act}$, the PRAE has a constraint to get the two encodings $z^{dsc}$ and $z^{act}$ closer such that descriptions and actions are combined. 
This allows the model to translate between descriptions and actions. 
We denote a batch of encodings of the descriptions as $\left\{z_k^{dsc} | 1 \leq k \leq K\right\}$ and encodings of the robot actions as $\left\{z_k^{act} | 1 \leq k \leq K\right\}$, where $K$ is the batch size. 
The binding loss $L_{shr}$ is defined as follows:
\begin{dmath}
    L_{\rm shr} = \sum _k^K \psi \left(z_{k}^{\rm act}, z_{k}^{\rm dsc} \right) + \sum_k^K \sum _{j\ne k} \max \left\lbrace 0,\\ \Delta + \psi \left(z_{k}^{\rm act}, z_{k}^{\rm dsc} \right) - \psi \left(z_{k}^{\rm act}, z_{j}^{\rm dsc} \right) \right\rbrace.
\end{dmath}
Here, $\psi$ is the Euclidean distance between the vectors. 
The first term brings the corresponding encodings closer and the second term pulls apart not correspond ones.
$\Delta$ represents the margin to enhance the loss. 
By minimizing the total loss of $L_{dsc}+L_{act}+L_{shr}$, the PRAE is able to translate bidirectionally between actions and descriptions.

\subsection{Transforming Word Embeddings Using a Retrofit Layer}

This study incorporates a retrofit layer between the pre-trained word embeddings and the input part of the description-RAE.
Our experiment used three fully connected layers as the retrofit layer.
Their activation functions are tanh.
This structure was chosen because it performs best without overfitting.
The input and output of the retrofit layer are word embeddings\footnote{
In order to focus on grounding by retrofitting, the present work was limited to a small dataset consisting of simple robot actions and linguistic descriptions with a fixed order.
For this reason, we did not use context-dependent embeddings, such as BERT~\cite{devlin-etal-2019-bert}; instead, we used the representative context-independent embeddings Word2Vec.}.
We trained this layer to output word embeddings that contain information on the robot’s actions and environments.

The retrofit layer is optimized by the same loss method described in the previous subsection 
and this layer is trained separately from the PRAE (Algorithm~\ref{alg:training}). 
This allows pre-trained word embeddings to be mapped in a non-linear manner, making it easier to generate actions 
while preventing the training process from overfitting.
Although the loss function aims to reconstruct each sequence of data across the available modalities, 
different inputs may result in highly similar outputs.
For example, an unseen word Y that is a synonym of a known word X may have a similar embedding and they should generate the same action.
When training two models concurrently, the model has too rich expressivity to map the pre-trained word embeddings and therefore may not be able to generate actions properly from unseen words.
By training two models separately, this expressivity is moderately constrained and the retrofit layer generates word embeddings that are appropriately mapped in a non-linear manner.
Thus, the retrofit layer can function as an amplifier that enhances the ``embodiment ingredient" latent in pre-trained word embeddings.

Our approach uses a small dataset to transform pre-trained word embeddings acquired purely from the text corpus into representations grounded to a robot's sensory-motor experience. 
In addition, the retrofit layer makes it possible for the robot to generate correct actions from descriptions, including unseen words, by transforming any unseen words to representations close to the trained ones.


\subsection{Training Procedure}
The detailed training procedure of the rPRAE is described in the Algorithm~\ref{alg:training}. 
To balance the two DNNs, we trained the retrofit layer and the PRAE multiple times alternately (Algorithm~\ref{alg:training}). 
This method prevents the training from overfitting, as it does with the 'chain-thaw' method~\cite{felbo2017using} that sequentially unfreezes and fine-tunes a single layer at a time. 
In addition, we partially train the PRAE in advance of the overall training process, which enabled the stabilization of the retrofit layer output.
Thus, this type of initialization could adequately determine the approximate output value of the retrofit layer. 

We optimized all the learnable parameters using the gradient descent method with a random batch sampled from the dataset. 
The hyperparameter $N$ is the maximum number of iterations, $n^{ini}$ is the number of iterations for the first train the PRAE, and $n^{ch}$ is the number of iterations to train the PRAE and the retrofit layer.

\floatstyle{spaceruled}
\restylefloat{algorithm}
\begin{algorithm}[bt] 
    \caption{Training procedures for the rPRAE}
    \label{alg:training}
    \begin{algorithmic} 
        \REQUIRE $X^{dsc}, X^{act}:$ paired dataset
        \REQUIRE $N, n^{ini}, n^{ch}:$ hyperparameters
        \STATE randomly initialize learnable parameters: $\theta_{AE}, \theta_{ret}$
        \FOR{$i=0$ to $N$}
            \STATE Sample a random batch $\left\{ x^{dsc}_i, x^{act}_i\right\}$ from $X^{dsc}, X^{act}$
            \STATE Calculate $L_{dsc}, L_{act}, L_{shr}$ by forward-path
            \STATE Compute total loss: $L_{all} \Leftarrow L_{dsc} + L_{act} + L_{shr}$
            \IF{($i < n^{ini}$) or ($int((i-n^{ini})/n^{ch})\%2=1$)}
                \STATE Update $\theta_{AE}$ by gradient method w.r.t. $L_{all}$
            \ELSE
                \STATE Update $\theta_{ret}$ by gradient method w.r.t. $L_{all}$
            \ENDIF
        \ENDFOR
    \end{algorithmic}
\end{algorithm}

%% file: subtex/3experiment.tex
\section{Experiments} \label{sec:experiment}

\begin{figure}[tb]
    \centering
    \includegraphics[width=8.6cm]{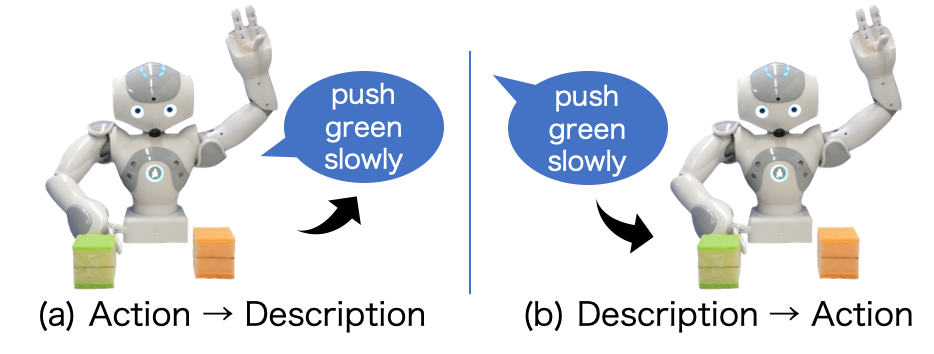}
    \caption{
        Overview of experiments with humanoid robot, NAO.
    }
    \label{fig:nao_exp}
\end{figure}

To evaluate the proposed method, we performed two learning experiments using a real humanoid robot, NAO: generation of a description from an action (Fig.~\ref{fig:nao_exp} (a), Experiment 1) and generation of an action from a description including unseen words (Fig.~\ref{fig:nao_exp} (b), Experiment2). 
In our experiments, we trained the rPRAE with a paired dataset of robot action sequences and word sequences. 
The robot action was obtained by generating predesigned action trajectories.
We used the dataset created by Yamada et al.~\cite{yamada2018paired}, as well as their experiment setup and data division, while extending the corresponding descriptions.

\begin{figure}[t]
    \centering
    \includegraphics[width=8.6cm]{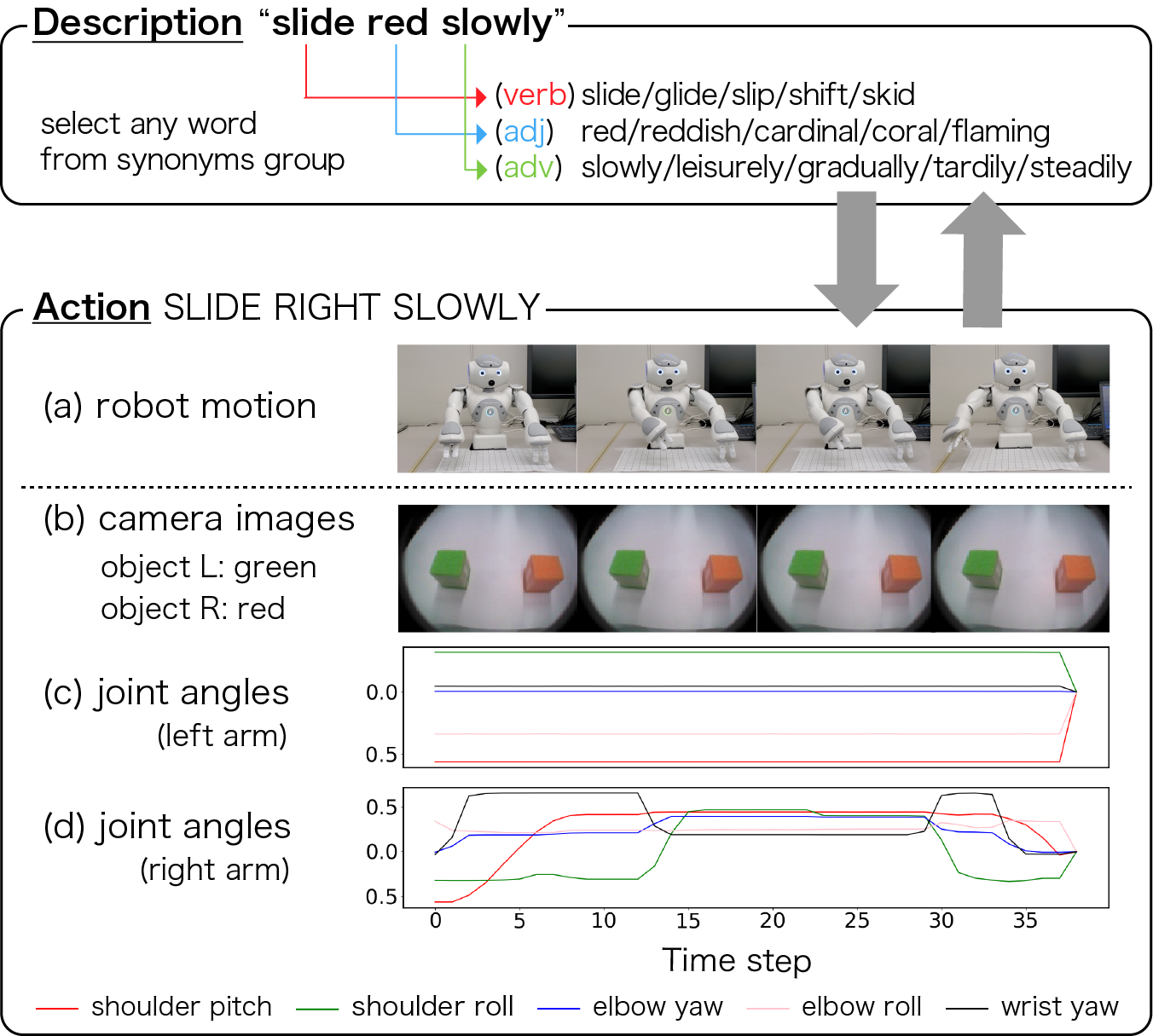}
    \caption{
        Example of paired sequences. The action is ``SLIDE-RIGHT-SLOWLY", description is ``\textbf{slide red slowly}", and red cube is placed on the right side of the robot. In the description data, we can select any word from each column to produce a same meaning description.
        In the action sequence, the pictures (a) show the robot motions. Pictures (b) are input images acquired from the robot's head camera. 
    }
    \label{fig:data_sample}
\end{figure}

\begin{table*}[tb]
    \vspace*{4pt}
    \centering
    \label{table:word_div} 
    \begin{tabular}{c|c c c|c c c|c c}
        \multicolumn{9}{c}{TABLE I: Word sets from divided synonym groups} \\
        \hline 
        \multicolumn{1}{c|}{} & \multicolumn{3}{c|}{verb} & \multicolumn{3}{c|}{adjective} &
        \multicolumn{2}{c}{adverb}\\
        \hline
        synonym group & \textbf{pull} & \textbf{push} & \textbf{slide} & \textbf{red} & \textbf{green} & \textbf{yellow} & \textbf{slowly} & \textbf{fast}\\
        \hline
        set1 & pull & push & slide & red & green & yellow & slowly & fast\\
        set2 & drag & shove & glide & reddish & greenish & yellowish & leisurely & speedily \\
        set3 & tug & thrust & slip & cardinal & olive & cream & gradually & swiftly\\
        set4 & yank & jostle & shift & coral & emerald & amber & tardily & rapidly \\
        set5 & lug & hustle & skid & flaming & chartreuse & tawny & steadily & quickly\\
        \hline
    \end{tabular}
    \vspace*{-4pt}
\end{table*}

\begin{table}[tb]
    \centering
    \label{table:data}
    \begin{tabular}{c|c|c}
        \multicolumn{3}{c}{TABLE II: Data division and the number of sequence pattern} \\
        \hline
        & trained description & unseen description \\
        \hline
        training actions & 3456 & 3294\\
        \hline
        test actions & 1152 & 1098\\
        \hline
    \end{tabular}
\end{table}

\subsection{Task Design}

\subsubsection{Robot Actions}

The task of the robot was to manipulate three cubes colored red, green, and yellow. 
Two cubes were placed in front of the robot, and ${}_3 \mathrm{P} _2 = 6$ cube arrangements were possible. 
To manipulate the cubes, the robot performed actions consisting of three components: motion, hand, and speed. 
Each component had two or three options; PULL / PUSH / SLIDE for the motion, LEFT / RIGHT for the hand, and SLOWLY / FAST for the speed. 
In this study, we utilized these options. 
For each cube arrangement, 18 types of actions were possible by considering combinations of these three components. 
Thus, the total number of possible action sequence patterns was 72, and each pattern was collected six times.

\subsubsection{Descriptions}

The descriptions, paired with the actions, were combinations of verbs (motion), adjectives (color), and adverbs (speed); \textbf{pull}/\textbf{push}/\textbf{slide} (verb) + \textbf{red}/\textbf{green}/\textbf{yellow} (adjective) +  \textbf{slowly}/\textbf{fast} (adverb).
For example, if you place a red object on the right and a green object on the left, 
the description corresponding to PUSH-RIGHT-SLOWLY is ``\textbf{push red slowly}."

In addition, we prepared a set of five synonyms for each word, called a ``synonym group."
In this paper, labels for synonym groups are shown in bold.
The total vocabulary consists of 41 words including five synonyms of the aforementioned 8 words and start/end symbols (BOS/EOS).

\subsection{Training Setup}

Figure~\ref{fig:data_sample} shows the sample paired sequence (action: ``SLIDE-RIGHT-SLOWLY" and description: ``\textbf{slide red slowly}").
In the description data (top row of Fig.~\ref{fig:data_sample}), we can select any words from the same synonym group to obtain a description with the same meaning.
In the action sequence (bottom of Fig.~\ref{fig:data_sample}), pictures (a) indicate the front view of the robot motion, and pictures (b) represent the input images acquired from the robot's head camera.
Pictures (c) and (d) indicate the joint angle values of the arms of the NAO.
Each robot action sequence consists of joint angles with 10 degrees of freedom (DoF) and 10-dimensional image features extracted by the CAE. 
In this study, the initial image was being input during an action sequence.
The robot action sequences were obtained in approximately 26 or 39 steps depending on the motion speed specified by the adverb. 
The values of joint angles were normalized from -0.8 to 0.8. 
The description data consisted of five elements, namely three words (forming a sentence) and BOS and EOS. 
Each element was a 300-dimensional vector acquired from pre-trained Word2Vec\footnote{\url{https:// code.google.com/archive/p/word2vec/}}~\cite{mikolov2013distributed}.

Each encoder and decoder of the PRAE was a one-layer biLSTM with 500 nodes. 
The dimensions of $z^{dsc}$ and $z^{act}$ were also 500. 
The margin $\Delta$ of the binding loss was 1.0. 
Each retrofit layer had 400 nodes, and the final output dimension was 300. 
We used the Adam optimizer (learning rate: 0.001), a batch size of 120, and a total of 100 training epochs. 
We trained 20736 sequences (described in Section ~\ref{sec:experiment_eval}); thus, one epoch contains 173 iterations, and the total number of iterations is $N=17300$. 
We set $n^{ini}=1$ and $n^{ch}=100$. 
These parameters were determined after considering several training settings.

\begin{table}[tb]
\label{table:act2desc} 
\begin{center}
    \begin{tabular}{c|c}
    \multicolumn{2}{c}{TABLE III: Success rates for the description generation.} \\
    \hline 
    training actions [\%] & test actions [\%] \\
    \hline 
    100.0 & 98.89\\
    \hline
    \end{tabular}
\end{center}
\end{table}

\begin{table*}[tb]
    \vspace*{4pt}
    \centering
    \label{table:desc2act_dtw} 
    \begin{threeparttable}
        \begingroup
        \begin{tabular}{c|c|c c c c}
            \multicolumn{6}{c}{TABLE IV: DTW score for generated action sequences} \\
            \hline
            \multicolumn{2}{c|}{} & \multicolumn{4}{c}{DTW [ - ] $\downarrow$}\\
            \hline
            \multicolumn{2}{c|}{The number of unseen words} & 0 & 1 & 2 & 3 (all)\\
            \hline 
            \hline
            training actions & rPRAE (ours) & 3.131 $\left(\pm{0.35587}\right)$ & 24.29 $\left(\pm{7.2385} \right)$ & 40.68 $\left(\pm{12.462} \right)$ & 56.76 $\left(\pm{16.037} \right)$\\
            (54 data)        & PRAE & 2.130 $\left(\pm{1.4167} \right)$ & 24.77 $\left(\pm{4.1224} \right)$ & 42.58 $\left(\pm{6.1349} \right)$ & 57.00 $\left(\pm{8.0764} \right)$\\
            \hline
            test actions     & rPRAE (ours) & 16.35 $\left(\pm{3.1388} \right)$ & 33.32 $\left(\pm{5.0976} \right)$ & 47.87 $\left(\pm{8.9638} \right)$ & 60.09  $\left(\pm{12.195} \right)$\\
            (18 data)        & PRAE & 17.17 $\left(\pm{3.5450} \right)$ & 34.97 $\left(\pm{3.7483} \right)$ & 49.46 $\left(\pm{5.5183} \right)$ & 60.74 $\left(\pm{6.2617} \right)$\\
            \hline
        \end{tabular}
        \endgroup
    \end{threeparttable}
    \vspace*{-4pt}
\end{table*}

\subsection{Data Division} \label{sec:experiment_eval}

We evaluated the trained model through five-fold cross-validation with respect to descriptions.
Each row of Table I indicates a word set used in a test in the cross-validation.
A word is selected from each synonym group, which is shown as a column, to form a description (e.g. ``pull cardinal quickly").
Cross-validation was performed on these five sets of words; four sets were used for the training, and the remaining set was used for the test. 
The words in the testing dataset were not included in the training dataset; thus, they were unseen words. 
In addition, we split the 72 action sequences into two parts: 54 sequences for the training dataset and 18 sequences for the testing dataset.
This division allowed us to test whether the model could generate appropriate descriptions or actions for novel combinations of robot actions and cube arrangements.

Table II summarizes the data division. 
Each cell of this table indicates the number of sequence patterns.  
The cells in the second row consist of the actions for the test, and the cells in the second column consist of the descriptions for the test that contain at least one or more unseen words. 
Because each action was collected six times, the number of sequences used is six times the number inferred from Table II. 
A total of $72$ (action) $\times 6 \times 5$ (verb) $\times 5$ (adjective) $\times 5$ (adverb) $= 54000$ sequences were possible, and we used $54 \times 6 \times 4 \times 4 \times 4 = 20736$ for training.
It should be noted that only the upper-left cell of Table II was used for training, and the remaining cells were used for testing.

\subsection{Experiments}

We evaluated the model's ability to generate linguistic descriptions and robot actions by conducting the following two experiments.

\textbf{Experiment 1: }
We confirmed that this model can generate descriptions from action sequences. 
We considered a description to be successfully generated if it satisfied the following two conditions: (1) the words are generated in the order of ``verb + adjective + adverb + EOS," and (2) each word belongs to the correct synonym group. 
The success rate was employed as a metric in Experiment 1.

\textbf{Experiment 2: }
To confirm that this model can generate robot actions from descriptions including unseen words, we evaluated the actions generated with and without the retrofit layer using the following three metrics. 
\begin{itemize}
    \item Metrics 1) The generated actions were quantitatively evaluated 
    using Dynamic Time Warping (DTW~\cite{muller2007dynamic}), which measures the distance between two sequences regardless of the differences in the sequence lengths.
    \item Metrics 2) The speed of an action specified by an adverb was evaluated using the total number of steps of the generated action. The generated action was considered successful if it took at most 30 steps for \textbf{fast} and more than 30 steps for \textbf{slowly}.
    \item Metrics 3) To evaluate the overall action, we investigated task achievement by the playback with the real robot.
    The generated action was considered successful if the robot moved to the correct side of its hand in the correct direction (PUSH/PULL/SLIDE) more than 1.8 cm at an appropriate speed. 
\end{itemize}

\begin{table}[tb]
    \centering
    \scalebox{0.96}[0.96]{
    \label{table:desc2act_fast} 
    \begin{threeparttable}
        \begingroup
        \begin{tabular}{c|c|c c c c}
            \multicolumn{6}{c}{TABLE V: Success rates for action generation in the speed condition} \\
            \hline 
            \multicolumn{2}{c|}{} &  \multicolumn{4}{c}{Success rates [\%]}\\
            \hline
            \multicolumn{2}{c|}{The number of unseen words} & 0 & 1 & 2 & 3 (all)\\
            \hline 
            \hline
            training actions & rPRAE (ours) & 100.0 & 97.78 & 95.56 & 93.33\\
            (54 data)        & PRAE & 100.0 & 98.56 & 96.77 & 95.43\\
            \hline
            test actions     & rPRAE (ours) & 100.0 & 97.78 & 95.56 & 93.33\\
            (18 data)        & PRAE & 100.0 & 98.43 & 96.23 & 95.19\\
            \hline
        \end{tabular}
        \endgroup
    \end{threeparttable}}
\end{table}

%% file: subtex/4result.tex
\begin{table*}[htb]
    \vspace*{4pt}
    \centering
    \begin{threeparttable}
        \begingroup
        \begin{tabular}{c|c|c|c c c|c c c|c}
            \multicolumn{10}{c}{TABLE VI: Success rates for task achievement using a robot} \\
            \hline
            \multicolumn{2}{c|}{} & \multicolumn{8}{c}{Success rates [\%]} \\
            \hline
            \multicolumn{2}{c|}{The number of unseen words} & \multicolumn{1}{c|}{0} &  \multicolumn{3}{c|}{1} & \multicolumn{3}{c|}{2} &  \multicolumn{1}{c}{3 (all)}\\
            \multicolumn{2}{c|}{unseen words} & - & verb & adj & adv & verb+adj & adj+adv & adv+verb & verb+adj+adv\\
            \hline 
            \hline
            training actions & rPRAE (ours) & 100.0 & 92.59 & 62.96 & 100.0 & 61.11 & 62.96 & 94.44 & 61.11\\
            (54 data) & PRAE & 100.0 & 83.33 & 48.15 & 100.0 & 37.04 & 48.15 & 85.19 & 31.48\\ 
            \hline
            test actions & rPRAE (ours) & 94.44 & 83.33 & 55.56 & 94.44 & 55.56 & 50.00 & 83.33 & 55.56\\
            (18 data) & PRAE & 88.89 & 72.22 & 27.78 & 88.89 & 27.78 & 27.78 & 61.11 & 27.78\\ 
            \hline
        \end{tabular}
        \endgroup
    \end{threeparttable}
    \label{table:nao} 
    \vspace*{-4pt}
\end{table*}

\begin{figure}[t]
    \centering
    \includegraphics[width=8.6cm]{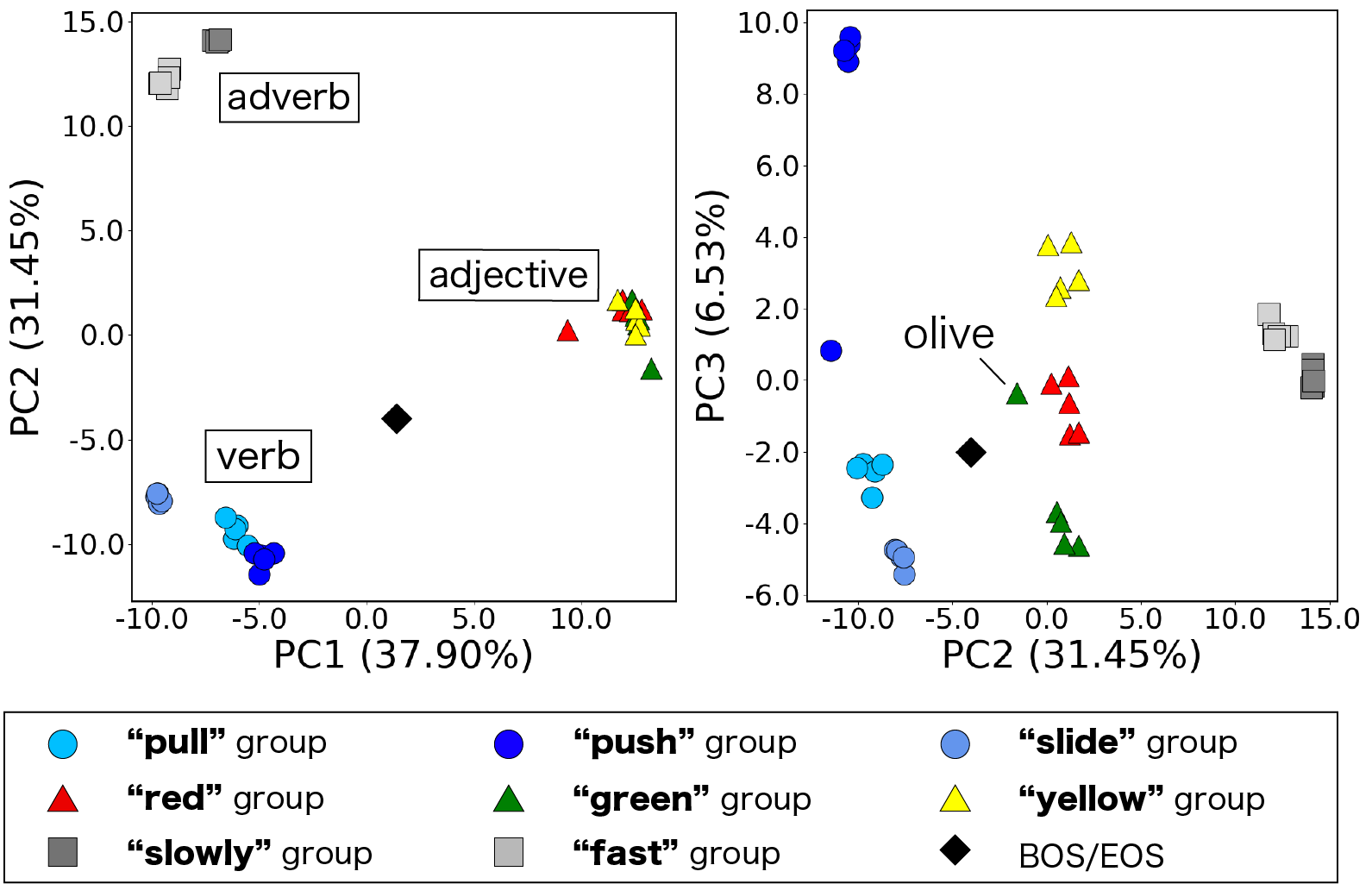}
    \caption{
        Visualization of retrofitted word embeddings using PCA.
    }
    \label{fig:pca}
\end{figure}

\begin{figure}[t]
    \centering
    \includegraphics[width=8.6cm]{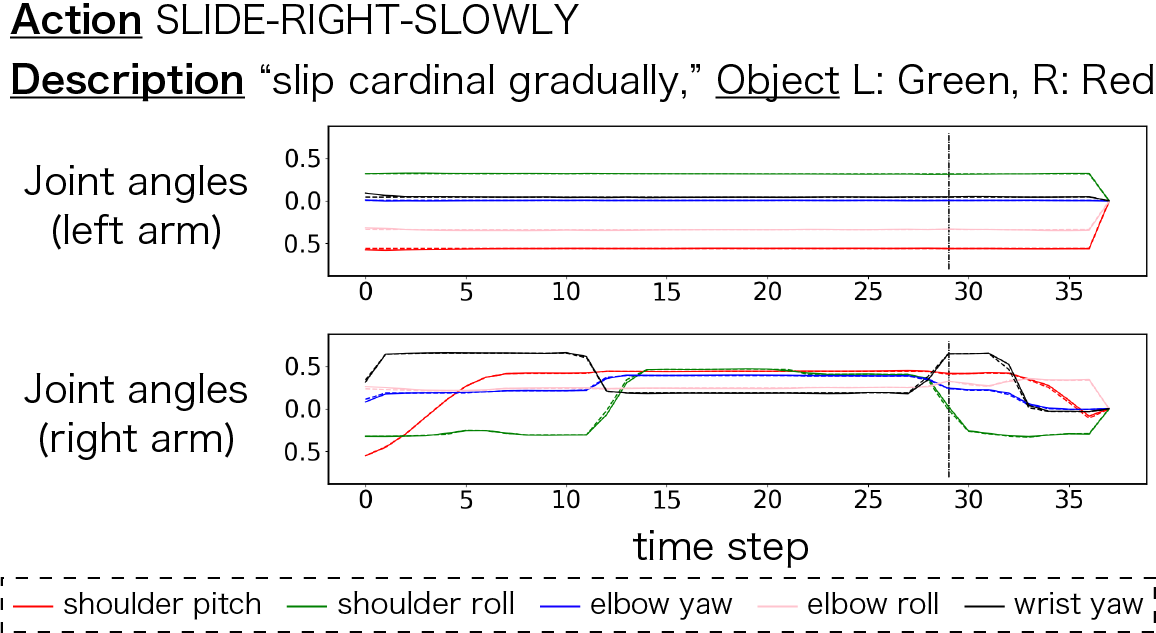}
    \caption{
        The action sequence generated by inputting the description ``slip cardinal gradually''.
        The solid lines indicate the generated joint angles. 
        The broken lines indicate the correct trajectories.
    }
    \label{fig:joint_result}
\end{figure}

\section{Results and Discussion} \label{sec:result}

\subsection{Generation of Linguistic Description from Action}

We evaluated the linguistic descriptions generated from the actions (Experiment 1). 
These descriptions were generated using the encoder of action-RAE and decoder of description-RAE. 
Table III shows the description generation success rates.
The success rate of the actions for the test was as high as $98.89\%$, indicating that our model could successfully translate correct descriptions even from untrained actions.
In other words, the model was capable of generating proper descriptions even for novel combinations of actions and environment.

\subsection{Generation of Action from Linguistic Description} \label{sec:desc2act}

We evaluated the actions generated from the linguistic descriptions (Experiment 2). 
We compared the rPRAE and the PRAE, which does not have the retrofit layer.

Metric 1) Table IV summarizes the results of the DTW score for the generated action sequences.
In all the cells of the table except for the top left cell, our proposed model showed better scores than those of the PRAE. 
These scores were particularly improved when the description contained one or two unseen words. 
On the other hand, the rPRAE presented issues regarding the stability of performance. 
The rPRAE failed to generate actions if the descriptions included particular unseen words that were not properly converted through the retrofit layer. 
As the success rate varies greatly depending on whether particular words are included in the testing dataset or not, the DTW scores of the rPRAE showed a high standard deviation. 
This may be a limitation of the simple structure of the retrofit layer that does not necessarily convert a word properly.

Metric 2) Table V presents the success rates for the motion speed condition.
Here, the rPRAE yielded worse scores than those of the PRAE.
The rPRAE failed only when a certain word ``rapidly" was included in the descriptions. 
It is assumed that this is because the pre-trained representation of ``rapidly" could not be retrofitted properly. 
Considering the results in Table IV and Table V, the generation performance of the rPRAE is noted to depend on the pre-trained embeddings.
However, this can be dealt with in various ways, such as via additional training before used in a real-world environment.  
Thus, the failures with respect to specific words that were improperly retrofitted are more tolerable than the unpredictable failures that would be attributed to the word combinations and unseen part-of-speech (POS) that the word 
would belong to.

\begin{figure}[tb]
    \centering
    \includegraphics[height=5.899cm]{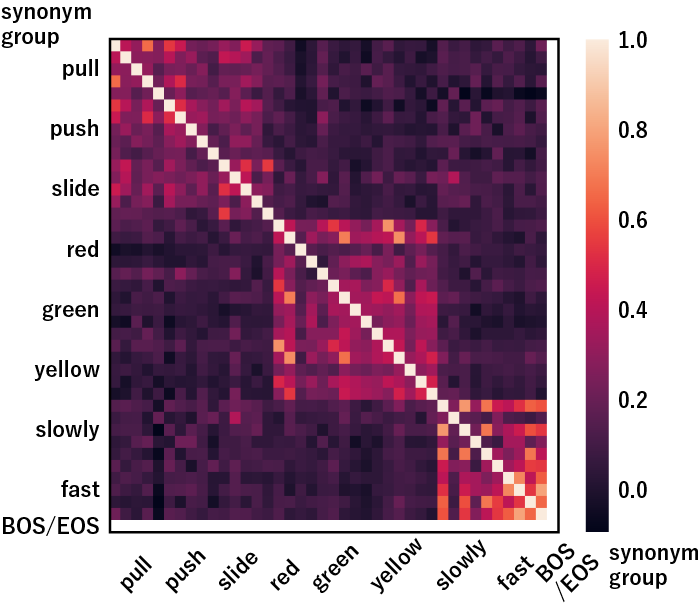}
    \caption{Cosine similarity of word embeddings by Word2Vec~\cite{mikolov2013distributed}.}
    \label{fig:cos_sim_origin}
\end{figure}
\begin{figure}[tb]
    \centering
    \includegraphics[height=5.899cm]{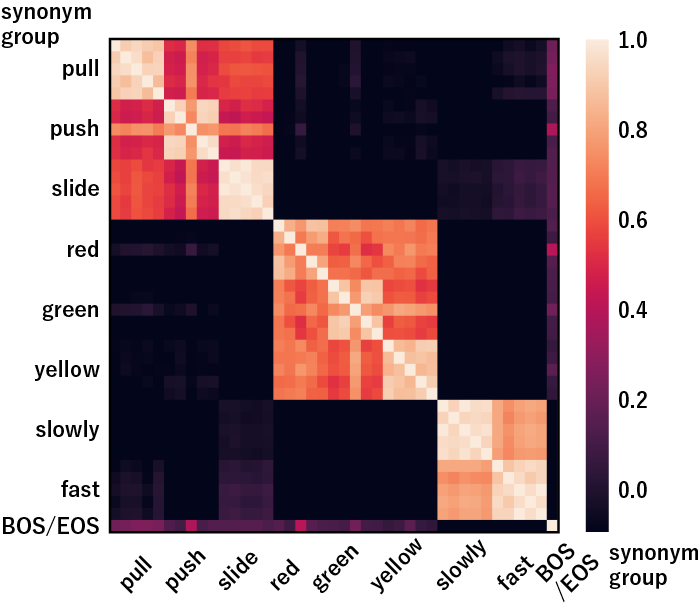}
    \caption{Cosine similarity of retrofitted word embeddings.}
    \label{fig:cos_sim}
\end{figure}

Metric 3) Table VI displays the success rates of task achievement using a real robot. 
The success rate of the PRAE was markedly worse than that of the rPRAE.
However, the rates of the rPRAE were similary low when the descriptions included unseen adjective words.
This is because the robot failed to generate actions when ``olive'' was included.
In this case, the robot generated the actions of \textbf{red}. 
To analyze the reason for this failure, we visualized retrofitted words via principal component analysis (PCA, Fig.~\ref{fig:pca}). 
In contrast to the case of the PC1-PC2 space, which indicates coherence of the POS, ``olive'' is placed separately from the synonym group in the PC2-PC3 space. 
According to WordNet~\cite{miller1995wordnet}, ``olive'' means ``of a yellow-green color similar to that of an unripe olive.''
Thus, affected by pre-trained word embeddings, ``olive'' had been embedded close to \textbf{red}, which is between \textbf{yellow} and \textbf{green}. 
Because the rPRAE relies on pre-trained word embeddings, the robot could not generate an action properly from the word that mixes multiple meanings.

Except for the above failures, our model generated proper actions without being affected by the number of unseen words in the descriptions and the POS they belonged to. 
Figure~\ref{fig:joint_result} shows a successful example of the generated action sequence. 
The generated joint angles (the solid lines) are apparently similar to the correct trajectories (the broken lines).
These results suggest that the retrofit layer transforms the pre-trained word embeddings, including unseen words, into a representation tailored to actions.

\subsection{Analysis of Retrofitted Word Representation}
To verify the effectiveness of the retrofit layer, we visualized word representations which was input to the description RAE. 
Figure~\ref{fig:cos_sim_origin} shows the cosine similarity of original word embeddings acquired by pre-trained Word2Vec~\cite{mikolov2013distributed} and Fig.~\ref{fig:cos_sim} shows the cosine similarity of word embeddings transformed by the retrofit layer.
Every five-row and -column block belongs to the synonym group, and the last row and column indicate BOS/EOS. 
The central row and column of each synonym group indicate the unseen word. 
The first three synonym groups are verbs, the next three groups are adjectives, and the last two groups are adverbs.

The results show that words are grouped according to the POS; moreover, the synonyms are grouped including unseen words. 
Unlike the original embeddings, even the antonyms \textbf{slowly} and \textbf{fast} are clearly separated.  
This grouping also can be seen in both embeddings in the visualization using PCA (Fig.~\ref{fig:pca} shows the retrofitted one).
We found that the rPRAE obtained word embeddings that could not be acquired under the distributional hypothesis.
Similar results may be confirmed in another context-independent word embeddings like GloVe~\cite{pennington2014glove}.
Although the retrofitted embeddings depends on the situation that the robot is exposed to, the proposed method is effective in terms of grounding pre-trained word embeddings learned from a text corpus to the real action and environment.

%% file: subtex/5conclusion.tex
\section{Conclusion}
We proposed a model to acquire integrated representations of unseen linguistic descriptions and sensory-motor experience. 
This model extends the bidirectional translation model of actions and descriptions by incorporating the retrofit layer. 
By training the retrofit layer and the bidirectional translation model alternately, the proposed model transforms pre-trained word embeddings to correspond to robot actions and the environment. 
The model performs bidirectional translation between robot actions and their linguistic descriptions even if they include unseen words. 
Experimental results show that the generation of robot actions is essentially independent of the number and order of unseen words in the description. 
In addition, the retrofitted word embeddings, including unseen words, were grouped according to the part-of-speech (POS) and synonym. 
These results indicate that the model acquires linguistic representations that cannot be obtained by performing training under the distributional hypothesis.

Future work should improve the model and extend the research setting by incorporating more complex actions and their complex descriptions.
This may consider complex descriptions involving a variety of temporal expressions like ``before/after," referential expressions, and polysemous words within/across the POS~\cite{juven:hal-02594725}\cite{7759133}\cite{hinaut:hal-02383530}.
To facilitate these developments, motion/video-captured datasets with a larger vocabulary and a powerful sentence encoder, such as BERT~\cite{devlin-etal-2019-bert}, are required.